\pdfoutput=1
\documentclass{bmvc2k}

\usepackage{dingbat}
\usepackage{multicol}
\usepackage{multirow}
\usepackage{wrapfig}
\usepackage{amsfonts}

\usepackage{caption}
\usepackage[export]{adjustbox}

\title{Deep Attentional Structured Representation Learning for Visual Recognition}
\runninghead{}{Deep Attentional Structured Representation Learning}

\def\etal{\emph{et al}\bmvaOneDot}

\addauthor{Krishna Kanth Nakka}{krishna.nakka@epfl.ch}{1}
\addauthor{Mathieu Salzmann}{mathieu.salzmann@epfl.ch}{1}

\addinstitution{
	Computer Vision Lab, EPFL\\
	Lausanne, Switzerland
}
\addinstitution{
	Computer Vision Lab, EPFL\\
	Switzerland
}

\def\etal{\emph{et al}\bmvaOneDot}

\newif\ifdraft
\draftfalse
\definecolor{orange}{rgb}{1,0.5,0}

\ifdraft
\newcommand{\comment}[1]{}
\newcommand{\ms}[1]{{\color{blue}{#1}}}
\newcommand{\MS}[1]{{\color{blue}{\bf #1}}}
\newcommand{\kn}[1]{{\color{red}{#1}}}
\newcommand{\KN}[1]{{\color{red}{\bf #1}}}

\else
\newcommand{\comment}[1]{}
\newcommand{\MS}[1]{}
\newcommand{\ms}[1]{ #1 }
\newcommand{\KN}[1]{}
\newcommand{\kn}[1]{ #1 }
\fi

\newcommand{\bB}{\mathbf{B}}
\newcommand{\bb}{\mathbf{b}}
\newcommand{\bX}{\mathbf{X}}
\newcommand{\bx}{\mathbf{x}}
\newcommand{\bv}{\mathbf{v}}
\newcommand{\bI}{\mathbf{I}}
\newcommand{\bW}{\mathbf{W}}
\newcommand{\bw}{\mathbf{w}}
\newcommand{\bH}{\mathbf{H}}
\newcommand{\bs}{\mathbf{s}}

\begin{document}
	
\maketitle
\begin{abstract}
Structured representations, such as Bags of Words, VLAD and Fisher Vectors, have proven highly effective to tackle complex visual recognition tasks. As such, they have recently been incorporated into deep architectures. However, while effective, the resulting deep structured representation learning strategies typically aggregate local features from the entire image, ignoring the fact that, in complex recognition tasks, some regions provide much more discriminative information than others.

In this paper, we introduce an attentional structured representation learning framework that incorporates an image-specific attention mechanism within the aggregation process. Our framework learns to predict jointly the image class label and an attention maps in an end-to-end fashion and without any other supervision than the target label. As evidenced by our experiments, this consistently outperforms attention-less structured representation learning and yields state-of-the-art results on standard scene recognition and fine-grained categorization benchmarks.

\comment{
Structured representations has proven highly effective to aggregate the local image descriptors by computing higher order statistics. Recent Convolutional Neural Networks, particularly NetVLAD, has modeled these representations as a learnable pooling layers showing great improvements. However, the individual VLAD encoded deep  CNN local feature vectors are typically aggregated uniformly, thereby not exploiting the discriminative power of certain semantic regions specific to the target label. 

In this work, we propose to model a discriminative attention-aware structured representation encoding by incorporating an image-specific attention mechanism. Importantly, we pose learning attention as an auxiliary task requiring no additional annotation. We incorporate the new soft-attentional weight into the structured layer and jointly optimize the complete system in an end-to-end manner.  Experiments show by integrating  attention mechanism to NetVLAD achieves state-of-art results on MIT-Indoor Scene dataset and competitive performance on three fine-grained datasets.
}
\end{abstract}

% !TEX root = bmvc_review.tex
% !TEX spellcheck = en-US

\section{Introduction}
\label{sec:intro}

In recent years, Convolutional Neural Networks (CNNs) have emerged as the de facto standard for visual recognition. Nevertheless, while they achieve tremendous success at classifying images containing iconic objects, their performance on more complex tasks, such as scene recognition and fine-grained categorization, remains comparatively underwhelming. This is partly due to their simple pooling schemes that fail to model the dependencies between local image regions. By contrast, in the realm of handcrafted features, structured representations, such as Bags of Words (BoW) ~\cite{bow3,BoW1,BoW2}, Vectors of Locally Aggregated Descriptors (VLAD)  ~\cite{vlad,allaboutvlad}  and Fisher Vectors (FV) ~\cite{fisher2010,fisher2013}, have been shown to be highly discriminative thanks to their aggregation of local information. As a consequence, they have started to re-emerge in the deep networks realm, with architectures such as NetVLAD~\cite{netvlad} and Deep FisherNet~\cite{deepfishernet}.

While effective for complex visual recognition tasks, these structured representations, whether based on handcrafted features or incorporated into deep networks, suffer from one drawback: They aggregate local information from the entire image, regardless of how relevant this information is to the recognition task. In practice, however, while certain image regions contain semantic information that contribute to the target label, others clearly don't. For example, in the image shown in Fig.~\ref{fig:arch}, from the MIT-Indoor dataset~\cite{mitindoor2009}, the region depicting washing machines gives us a much stronger cue of the class \emph{laundry} than the regions containing  the person and the background. Incorporating information from these latter two regions, which can appear in many other scene categories, will typically yield less discriminative image representations.

\begin{figure}
	\centering
	%\vspace{-0.3in}
	\includegraphics[width=0.95\linewidth, height=3.5cm]{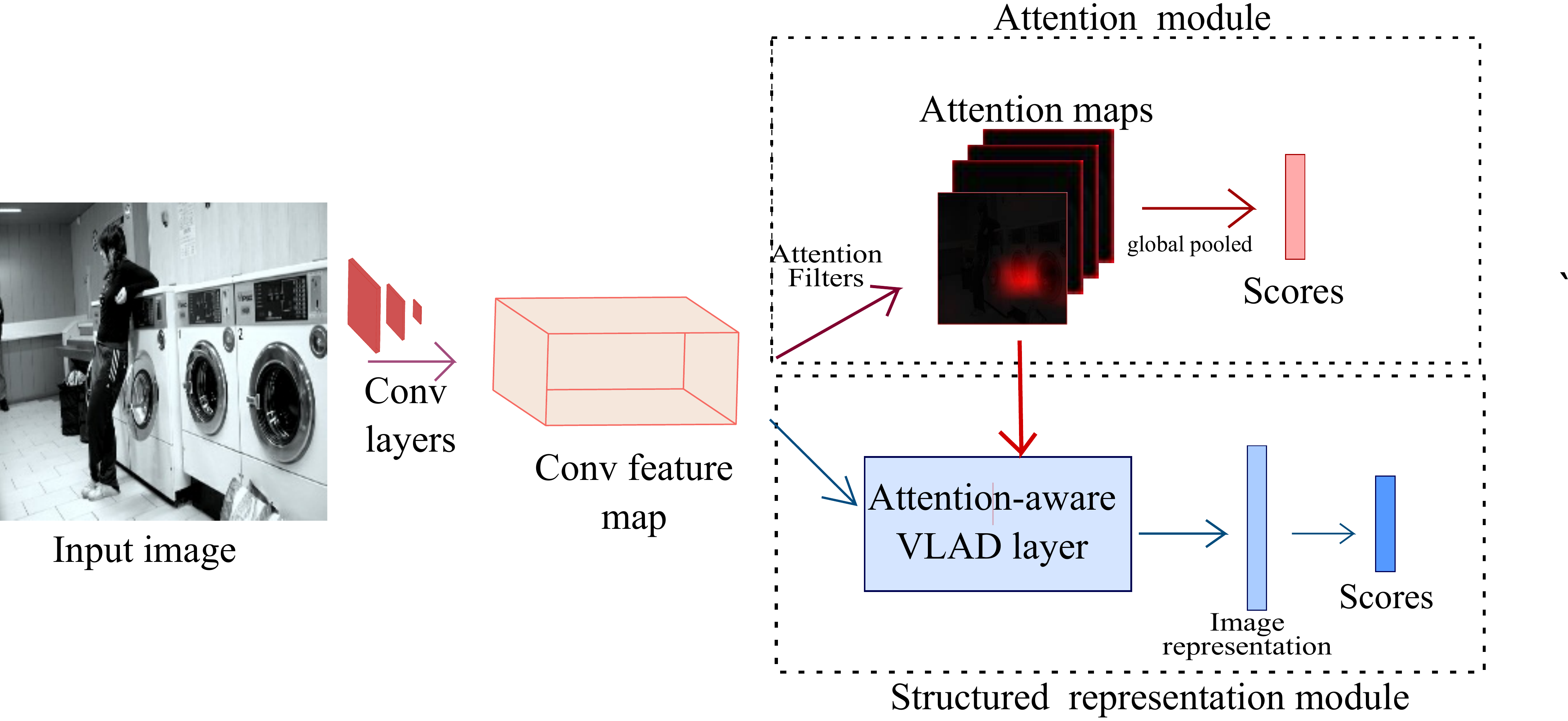}
	\vspace{0.12in}
	
	\caption{\label{fig:arch} \textbf{Attentional structured representation network.} Our network consists of two branches with a shared base feature extraction CNN. The attention module produces class-specific attention maps, which are then incorporated into the VLAD module that outputs an attention-aware VLAD representation. Note that, while we focus on the VLAD case here, as evidenced by our experiments, our approach applies to any structured representation.}
	\vspace{-0.2in}
	%{-0.2in}
\end{figure}

In this paper, we address this by introducing a novel deep attentional structured representation network for visual recognition. Our network incorporates an image-specific attention mechanism that encourages the learnt structured representation to focus on the discriminative regions of the image. We then learn to predict jointly the input image class label and the spatial attention map without requiring any annotations for the latter. 

Our framework is depicted by Fig.~\ref{fig:arch} for the case of a VLAD aggregation strategy. It consists of two streams that share a base network extracting deep features: The attention module and the VLAD module. The attention module, based on the framework of~\cite{Attpool}, learns a set of filters that transform the deep features into $C$ heatmaps encoding attention for the $C$ classes of interest. The VLAD module then exploits these heatmaps to form an attention-aware VLAD vector from the deep features of the base network. We train our network with a combination of two losses that encourage both the attention maps and the final attention-aware VLAD representation to be discriminative. Note that, while Fig.~\ref{fig:arch} focuses on the VLAD case, as evidenced by our experiments, our approach generalizes to any local feature aggregation strategy.

In short, we contribute the first systematic integration of an attention mechanism within a structured image representation learning framework. We demonstrate the benefits of our approach on four challenging visual recognition tasks, including scene recognition on MIT-Indoor~\cite{mitindoor2009} and fine-grained categorization on the UCSD Birds~\cite{ucsdbirds}, FGVC Aircrafts~\cite{aircraft} and Stanford Cars~\cite{stanfordcars} datasets. Our attentional structured representation learning strategy consistently outperforms its standard attention-less counterpart and yields state-of-the-art results on several of the above-mentioned datasets.

\comment{
In this work, we propose a novel deep attentional structured representation network for visual recognition tasks. We incorporate an image-specific attention  into structured layers with focus to discriminate regions contributing to the target label. We pose learning spatial-attention map as an auxiliary task along with problem at hand without requiring additional annotation.  The attention module based on the framework of Rohit \etal ~\cite{Attpool} is seamlessly integrated into the existing system. The key idea behind this module is to learn a set of filters which transforms input feature map into $N$ output maps(for $N$ class dataset) of same spatial size, which are global pooled to minimize softmax loss over the target label. In this regard, we share the base network for the two tasks, and jointly learn the structured representation and attention map in an end-to-end manner. We re-visit the VLAD encoding scheme and combine an additional soft-attentional weight into the residual vector term. In this way, the resultant residual vectors at each cluster orients in the direction of descriptors with high attention.    
 
We follow a 2-step training procedure to achieve this. In the first step, we finetune the base network with attention module to estimate regions  contributing to the target label, and in second step, we train the whole attention-aware system which is additionally regularized with attention loss. Our Contributions can be summarized as follows: a. We instigate the first systematic integration of attention mechanism in conjunction with structured image representations b.  We propose to learn the spatial attention as an auxiliary task to discover the discriminative regions contributing to target label which is incorporated into the structured layers c. We conduct experiments on MIT-Indoor scene dataset achieving the state-of-the-art performance and competitive results on three fine-grained datasets (UCSD Birds, FGVC Aircraft, Stanford Cars) in comparison to standard VLAD. 
}

%
%
%The remainder of this paper is organized as follows. First, we provide an overview of related work in Section~\ref{sec:RelatedWork}. We then describe our deep attentional architecture in Section~\ref{sec:Method} and present and analyze our experimental results in Sections~\ref{sec:ExperimentalResults} and~\ref{sec:Conclusion}.

% !TEX root = bmvc_review.tex
% !TEX spellcheck = en-US

\section{Related Work}
\label{sec:RelatedWork}

%\textbf{Structured Representations}: 
Over the years, visual recognition has attracted a huge amount of attention in Computer Vision. Before the deep revolution in 2012, most methods adopted a two step pipeline consisting of extracting handcrafted features and training a classifier, such as Support Vector Machines~\cite{svm1998} or Boosting ~\cite{boosting1996}. In this pipeline, the core Computer Vision research was targeted towards extracting discriminative image features. In particular, Bags of Visual Words (BoW)~\cite{BoW1,BoW2,bow3},  based on local features such as SIFT~\cite{sift} or BRIEF~\cite{brief}, have proven effective for image recognition. Later, such histogram-based features were extended to VLAD~\cite{vlad} and Fisher Vectors~\cite{fisher2010,fisher2013}, which model higher-order statistics of the data w.r.t. the codewords. After the remarkable performance of AlexNet~\cite{Alexnet}, much of the visual recognition research turned to deep learning strategies. While many new architectures do not explicitly focus on extracting structured representations, some work has nonetheless attempted to leverage the lessons learnt from handcrafted features. In particular,~\cite{vladmultiscale} performs multi-scale orderless pooling of deep CNN features, and~\cite{semanticFV,FVFC} compute Fisher encodings of similar deep features. In contrast with these approaches that still separate feature extraction from classifier learning, NetVLAD~\cite{netvlad} and Deep FisherNet~\cite{deepfishernet} constitute the first attempts at introducing learnable VLAD and Fisher Vector layers, respectively, within an end-to-end learning formalism. More recently,~\cite{mfafvnet} proposed to make use of a mixture of factor analyzers to model an accurate Fisher Vector with full covariance matrix. While the previous methods all rely on histogram-based descriptors, in the context of fine-grained categorization and texture recognition, other structured representations, in the form of covariance matrices have been used~\cite{bcnn,Secondorderkaicheng2018,improvedbcnn}. In any event, all these methods, whether using hand-crafted features or relying on deep learning, aggregate local features from the entire image, without accounting for the fact that only parts of the image contain information that contributes to the target class label. This will typically reduce the discriminative power of the resulting representations.

For complex tasks, such as scene classification and fine-grained categorization, some research has nonetheless attempted to focus the feature extraction process on discriminative image regions. In the context of scene recognition, this was achieved by modeling the scene with mid- (or high-)level representations~\cite{objectbank2010}, such as detected semantic visual attributes~\cite{sceneattri}, patch-based codewords obtained via discriminative clustering~\cite{unsupervisedpatch} and object-oriented representations learnt from a manually-created database of typical scene objects~\cite{metaobjectsscene2015}. For fine-grained categorization, several works exploit bounding box annotations to learn part detectors~\cite{partrcnn,spda}. The use of such additional annotations was then removed in~\cite{twolevelattention}, which learns part templates by clustering  deep features. More recently,~\cite{RACNN2017,MACNN2017} introduced end-to-end learning strategies to automatically identify discriminative regions for fine-grained recognition. 

\comment{
Recent works~\cite{sceneattri,unsupervisedpatch,metaobjectsscene2015,objectbank2010} in scene classification focused on modeling scene using mid-level elements for learning representations. This include using semantic visual attributes  ~\cite{sceneattri}  based detectors to extract features for classification. The authors in  ~\cite{unsupervisedpatch} performed discriminative clustering on image patches in feature space to find most representative parts and used as codebook for feature encoding . Later,  ~\cite{metaobjectsscene2015}  manually created an database of  objects typically present in scene datasets to learn  object-oriented representations. On  the fine-grained datasets, a few works attempt to investigate the effectiveness of exploiting features from different convolutional layers [~\cite{crossconv2015}, ~\cite{alllayers}]. Some more include~\cite{FCN2015} using the features from bottom convolutional layers in the top layers to extract richer representation for the task of segmentation. While the works in~\cite{partrcnn,spda}  use annotations of bounding boxes to learn part-detectors for fine-grained recognition, which might not be optimal for scaling to large datasets withoout annotations. Later works by~\cite{twolevelattention} includes an unsupervised approach to learn part templates by clustering deep CNN features. More recently, focus has been to study to automatically locate discriminative regions recurrently for fine-grained tasks~\cite{RACNN2017,MACNN2017}, however this are tailormade for fine-grained but extendable to other datasets.
}

The above-mentioned works typically reason about the notion of parts, or objects in a scene. In the rare cases that don't require part annotations during training~\cite{RACNN2017,MACNN2017}, the input image is first processed globally to identify regions with high attention, which are then cropped into multiple parts that are processed individually. By contrast, our network processes the input image in a single forward pass, without explicitly relying on the notion of parts.  
In essence, these methods are therefore tailored to the specific problem they tackle. By contrast, here, we exploit the more general notion of visual attention and produce heatmaps encompassing the discriminative regions in the image. This does therefore not require any prior knowledge about the data at hand. Our formalism builds upon the attention framework of~\cite{Attpool}, but with the additional goal of leveraging structured representations. As a consequence, and as evidenced by our results, our approach yields higher accuracies than \ms{both attention-less methods and unstructured attentional pooling} in all the tasks we tested it on.

% !TEX root = bmvc_review.tex
% !TEX spellcheck = en-US

\section{Method}
\label{sec:Method}
In this section, we introduce our novel attentional structured representation learning framework depicted by Fig.~\ref{fig:arch}.
We first present the structured representation and attention modules, and finally our approach to integrating them in an end-to-end learning formalism.

\subsection{Structured Representation Module}
\label{sec:structured}

Structured representations aggregate local descriptors into a global feature vector of fixed size using a visual codebook. In particular, here, we focus on VLAD, which has proven highly effective. As will be evidenced by our experiments, however, our framework generalizes to other aggregation strategies.

In contrast to BoW that only store information about which codeword each local descriptor is assigned to, VLAD also computes the residual distance of the descriptor to the codeword. 
%BoW model is a simplified version of VLAD,  storing only the frequency of visual words.
%This pooling schemes are extended to deep networks as learnable layers replacing hard quantization with a soft-assignment policy.
To incorporate this into a deep learning framework, the hard codeword assignment of each descriptor is replaced by a soft one.
More specifically, let $\bI$ be an image input to a CNN, and $\bX \in \mathbb{R}^{W \times H \times D}$ the feature map output by the last convolutional layer, with spatial resolution $W \times H$ and $D$ channels. $\bX$ can then be thought of as $N = W \times H$ local descriptors $\bx_{i}$ of dimension $D$. Given a codebook $\bB$ with $K$ codewords, VLAD produces a $DK$-dimensional representation of the form
\begin{equation}
    \bv = [ \bv_0^T, \bv_1^T, \cdots, \bv_K^T]^T\;,
    \label{eq:vlad_vector}
\end{equation}
where $\bv_k \in \mathbb{R}^D$ is given by
\begin{equation}
       \bv_k = \sum_{i=1}^N a_k(\bx_{i})  \left( \bx_{i} - \bb_k \right)\;,
       \label{eq:vlad_residual}
\end{equation}
with $\bb_k$ the $k$-th codeword of codebook $\bB$. The values $a_k(\bx_{i})$ represent the assignment of descriptor $\bx_{i}$ to codeword $\bb_k$. In the standard VLAD formalism, these assignments are binary, with each descriptor being assigned to a single codeword. Within a deep learning context, for differentiability, these assignments can be relaxed and expressed as
\begin{equation}
      a_k(\bx_{i})= \frac{ e^{-\alpha\left\| \bx_{i}-\bb_k  \right\|^2}}{\sum_{k'}{e^{-\alpha\left\| \bx_{i}-\bb_{k'} \right \|^2}}}\;,
	\label{eq:vlad_soft_assgn}
\end{equation}	
with $\alpha$ a hyperparameter defining the softness of the assignments.

The resulting VLAD vector then acts as input to the classification layer of the deep network. While effective, as discussed above, the VLAD representation aggregates information from the entire image, regardless of whether the local descriptors correspond to discriminative regions or not. Below, we first discuss a general attention module, which is able to identify relevant image regions, and then introduce our approach to incorporating this information within our structured representation learning formalism.

\subsection{Attention Module}
\label{sec:attention}

It has been shown multiple times that CNNs were not only effective at predicting the class label of an image, but could also localize the image regions relevant to this label~\cite{CAM,CWCAM,gradCAM}. Most existing approaches to performing such a localization, however, work as a post-training step. By contrast, our attention module, based on the framework of~\cite{Attpool}, produces attention maps that are actively used during training. Furthermore, it combines top-down attention, modeling class-specific information, with bottom-up attention, modeling class-agnostic information, or, in other words, a form of image saliency.

Specifically, let $\bX$ be the same final $W \times H \times D$ convolutional feature map as in Section~\ref{sec:structured}. Our attention module consists of an additional $1 \times 1$ convolutional layer with one class-agnostic filter with parameters $\bw_{ca} \in \mathbb{R}^{D \times 1}$ and $C$ class-specific filters whose parameters can be grouped in a matrix $\bW_{cs} \in \mathbb{R}^{D \times C}$, where $C$ is the number of classes of the problem at hand. This convolutional layer produces a class-agnostic heatmap $\bH_{ca}$  and class-specific heatmaps $(\bH_{cs}^{1},\cdots,\bH_{cs}^{C})$, each of spatial resolution $W \times H$. Each class-specific heatmap is then multiplied element-wise by the class-agnostic one, yielding $C$ attention maps $(\bH^{1},\cdots,\bH^{C})$.

Training the attention module can be achieved by global average pooling of each of the attention maps, which produces a score for each class. These scores are then passed through a softmax layer, and the resulting probabilities $\{p_c\}$ used in a standard cross-entropy loss
\begin{equation}
L_{att} = - \frac{1}{S} \sum_{s=1}^S \log(p_{c^*}(\bI_s))\;,
\label{eq:att_loss}
\end{equation}
where $S$ is the number of samples in a mini-batch and $p_{c^*}(\bI_s)$ is the probability of the ground-truth class for sample $s$. This was the procedure used in~\cite{Attpool} to train an attentional deep network. Below, we propose to rather make use of the attention maps to further build a more discriminative structured representation. As evidenced by our results, this allows us to achieve consistently higher recognition accuracies.

\subsection{Attention-aware Feature Aggregation}

Our goal is to make use of the attention maps when aggregating the local descriptors into a structured representation. To this end, instead of global average pooling the maps, we generate a single attention map, \ms{which can be interpreted as a weight $w(\bx_{i})$ for every descriptor $\bx_{i}$, and is defined as
\begin{equation}
w(\bx_{i})= \frac{ \max\limits_{l}     \bH_{i}^{l}}{\sum\limits_{i'}\max\limits_{l}     \bH_{i'}^{l}}\;,
\label{eq:soft_att}
\end{equation}
where $\bH_{i}^{l}$ indicates the attention-weight corresponding to feature $\bx_{i}$ in the attention map of class $l$ from Section~\ref{sec:attention}. The resulting attention map has the same spatial resolution as the final deep feature map. We then use it to re-weight the aggregation scheme of Eq.~\ref{eq:vlad_residual}.
Specifically, we re-write Eq.~\ref{eq:vlad_residual} as
\begin{equation}
\bv_k = \sum_{i=1}^N w(\bx_{i}) a_k(\bx_{i})  \left( \bx_{i} - \bb_k \right)\;.
\label{eq:vlad_att}
\end{equation}
Following common practice~\cite{vlad,netvlad}, we perform $L_2$ normalization of each $\bv_k$ to remove burstiness, followed by a final $L_2$ normalization of the entire vector $\bv$. The resulting representation is then passed to a classification layer.
}

\comment{
 $\bH_{R}$ through Eq.~\ref{eq:soft_att} that has the same spatial resolution as the final deep feature map. We then use the resulting attention map  \kn{$\bH_{R}$} to re-weight the aggregation scheme of Eq.~\ref{eq:vlad_residual}.
Specifically, we re-write Eq.~\ref{eq:vlad_residual} as
\begin{equation}
\bv_k = \sum_{i=1}^N w(\bx_{i}) a_k(\bx_{i})  \left( \bx_{i} - \bb_k \right)\;,
\label{eq:vlad_att}
\end{equation}
where the weights $w(\bx_{i})$ are obtained from the attention maps of Section~\ref{sec:attention} as
\begin{equation}
w(\bx_{i})= \frac{ \max\limits_{l}     \bH_{i}^{l}}{\sum\limits_{i'}\max\limits_{l}     \bH_{i'}^{l}}\;,
\label{eq:soft_att}
\end{equation}
where $\bH_{i}^{l}$ indicates the  \comment{removed the term: location} \kn{attention-weight} corresponding to feature $\bx_{i}$ in the attention map of class $l$.
%We use the normalized form of this term  and incorporate into residual vector calculation.
Following common practice~\cite{vlad,netvlad}, we perform $L_2$ normalization of each $\bv_k$ to remove burstiness, followed by a final $L_2$ normalization of the entire vector $\bv$. The resulting representation is then passed to a classification layer.
}

%\subsection{Joint Attentional Structured Representation Learning}

Ultimately, our network combines an attention module with a structured representation learning one. Both modules share the base network up to the final convolutional feature map. To train our network, we first pre-train the base network with the attention module only using $L_{att}$ from Eq.~\ref{eq:att_loss}. We then continue training the entire network in an end-to-end manner by minimizing a loss of the form
\begin{equation}
L = L_{cls} + \lambda L_{att}\;, 
\label{eq:joint_loss}
\end{equation}
where $L_{cls}$ is a cross-entropy loss on the output of the classifier acting on the structured representation $\bv$, and $\lambda$ is a hyper-parameter setting the relative influence of both terms. At test time, we then take the prediction from the VLAD branch of the network.

Note that training is not only done w.r.t. the network parameters, but also w.r.t. to the codebook $\bB$. As suggested in~\cite{netvlad}, and motivated by~\cite{allaboutvlad} to adapt VLAD descriptors to new datasets, we decouple the soft assignment $a_k(\bx_{i})$ from the codeword $\bb_k$. That is, we re-write the assignment  $a_k(\bx_{i})$ of Eq.~\eqref{eq:vlad_soft_assgn} as
\begin{equation}
		 a_k(\bx_{i})= \frac{ e^{\bs_{k}^T\bx_{i}+h_k }}{\sum_{k'}{e^{\bs_{k'}^T\bx_{i}+h_{k'}}}}\;,
			\label{eq:vlad_soft_assgn_decoupled}
\end{equation}	
where $h_{k} = -\alpha \left \| \bb_k \right\|^2$ and $s_k = 2\alpha\bb_k$ are treated as independent parameters. 
%The three sets of parameters $\{\bb_k\}, \{\bs_k\}, \{h_k$\} are treated independent when training the network

 \subsubsection{Geometric Interpretation of Attention}\label{sec:norm}
 \begin{wrapfigure}{r}{0.4\textwidth}
 	\vspace{-0.2in}
 	\centering
 	\includegraphics[valign=T, width=5cm]{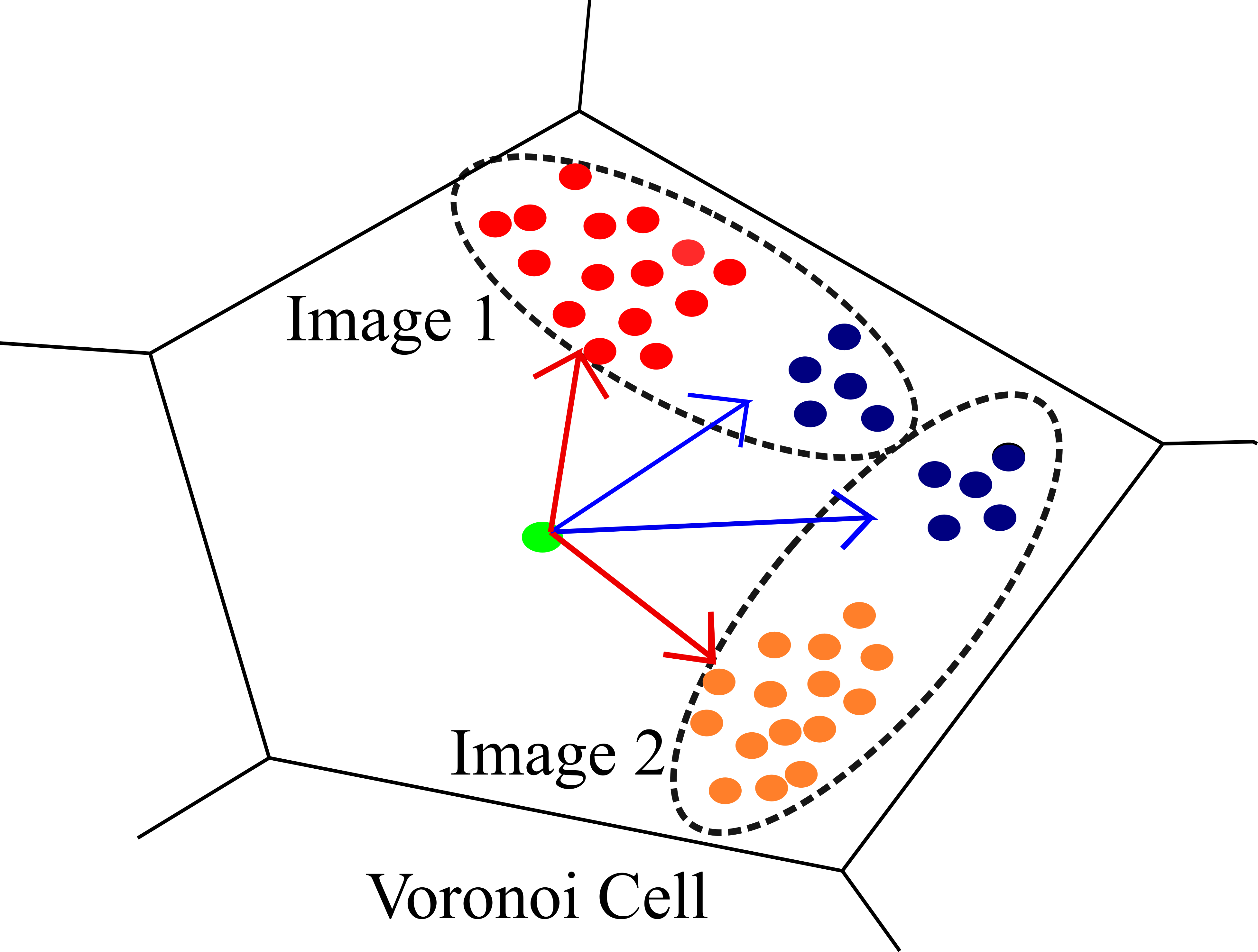}
 	\vspace{0.1in}
 	\caption{\label{fig:geometric_meaning} Geometric interpretation of attention}
 	\vspace{-0.1in}
 \end{wrapfigure}

\paragraph{}Consider the features of two images from same class with different backgrounds that are assigned to the same codeword, depicted as a Voronoi cell in Fig.~\ref{fig:geometric_meaning}. The features with high attention are shown in blue and those with low attention in red and orange, respectively. While ignoring attention would yield residual vectors pointing in almost opposite directions, our attention-aware aggregation produces vectors with high cosine similarity, shown as blue arrows. The inverse reasoning can be made for images from two different classes but containing common elements that are irrelevant to the class labels: By ignoring attention, these shared elements would yield components with high cosine similarity, thus decreasing the discriminative power of the complete VLAD vector. Attention allows us to discard these shared elements.

% !TEX root = bmvc_review.tex
% !TEX spellcheck = en-US

\section{Experiments}
\label{sec:ExperimentalResults}

We first present the datasets used in our experiments and implementation details for our model. Then, we demonstrate the benefits of our attention-aware structured representation learning framework over its attention-less counterpart and over unstructured attentional pooling, and finally compare our results to the state of the art on each dataset. 
\ms{We provide additional results and ablation studies in the supplementary material.}

\comment{
We report the accuracy of standalone attention module without structured representation in Section~\ref{sec:Attention_results}. The results suggest attention module can be reliably used as weakly-supervised localization of informative regions contributing to target label. 
We then show the benefits of attention-aware structured representation learning framework in Section~\ref{sec:Att_structured} consistently outperforming the attention-less systems.
%We further conduct ablation studies on the effect of regularization parameter on the model in Section~\ref{sec:lambda_effects}.
We finally compare our method with current state-of-the-art in Section~\ref{sec:state-of-the-art}, particularity with methods that use higher order statistics~\cite{bcnn,mfafvnet,netvlad,fisher2013}
}

\subsection{Datasets}
\label{sec:Datasets}

We experiment on the MIT-Indoor scene recognition dataset and on three fine-grained categorization datasets, namely CUB-200, Stanford cars and aircraft. We discard the part annotations but conduct experiments with and without bounding box annotations on fine-grained datasets.

\textbf{MIT-Indoor} is a widely used benchmark dataset for scene classification with 67 classes.
%The dataset contains scene layouts with large intra-class variance and high inter-class similarity. 
We use the train/test split of~\cite{mitindoor2009} consisting of roughly $80$ training and $20$ test images.

\textbf{CUB-200}  is a challenging dataset with $11,788$ images of $200$ bird species, with an average of $60$ images per class. The dataset has extremely large variations in pose, size and viewpoints. We use the standard train/test split of~\cite{ucsdbirds}. 
%We discard the part annotations but conduct experiments with and without bounding box annotations.

\textbf{FGVC-Aircraft} contains $100$ different aircraft models with roughly $100$ images for each model. We adopt the same train/test split as in~\cite{aircraft}. 
%We do not use any box annotations for this dataset.

\textbf{Stanford Cars} is a $196$ class dataset~\cite{stanfordcars} of $8144$ training images and $8041$ test images. Heavy background clutter makes this dataset challenging. %We carry out experiments both with and without box annotations.

\subsection{Implementation Details}
\label{sec:implment_details}
We use the VGG-16~\cite{vgg16} model pre-trained on Imagenet~\cite{Alexnet} as our base model and that of the baselines, with the $conv5\_3$ features before $ReLU$ activation as final convolutional features for aggregation. 
%We use same settings as baselines for fair comparisons. 
Following prior work~\cite{mfafvnet,bcnn}, we resize the images to $512 \times 512$ for MIT-Indoor,  and $448 \times 448$ for the fine-grained datasets.\comment{Data augmentation is carried out on all datasets by performing random cropping and horizontal flipping.} \ms{At test time, we flip the image and average the predictions for the original and flipped image}. For structured representations, we fix the codebook size to $K=64$ for VLAD and $K=4096$ for the BoW experiments. We initialize the weights  %\{$\bs_{k}$, $h_k$, $b_k$\}
of the VLAD layer with $K$-means clustering of the $conv5\_3$ features.
We set $\alpha$ in Eq.~\ref{eq:vlad_soft_assgn} to 100, and $\lambda$ in Eq.~\ref{eq:joint_loss} to 0.4. 

\textbf{Training:} We use the ADAM optimizer~\cite{adam} with parameter $\epsilon=10^{-4}$, batch size of $16$ and a weight decay of $0.0005$ for all experiments. \comment{We accumulate the gradients for a batch size of $32$ before updating the network weights.} 
%Furthermore, to avoid over-fitting, we use dropout with a probability of $0.5$ after the structured representation layer. 
\comment{For fine-grained datasets, we first pre-train the attention network with a learning rate $\eta=0.0001$, and then continue training the entire network with a higher $\eta=0.01$ for the classification layer and $\eta=0.0001$ for the other layers.} We first pre-train the attention network with $\eta=0.0001$ for 20 epochs. \ms{For scene recognition, we then train the classification layer with $\eta=0.01$ for $5$ epochs, and further train the layers above $conv5$ with $\eta=0.00001$ for $25$ epochs. For the fine-grained datasets, we train with $\eta=0.01$ for the classification layer and $\eta=0.0001$ for the layers above $conv5$ for $50$ epochs, with a decay rate of $0.1$ every $15$ epochs.} 
\comment{Further, a. we train with $\eta=0.01$ for the classification layer and $\eta=0.0001$ for the  layers above $conv5$ for $50$ epochs and a decay rate of $0.1$ for every $15$ epochs on fine-grained datasets  
b. while for scene dataset, we train the classification layer  with $\eta=0.01$ for $5$ epochs and further train layers above $conv5$ with uniform  $\eta=0.00001$ for $25$ epochs.} 
%\KN{(this is best configuration for individual datasets)}
\subsection{Results}
\label{sec:Results}
We first compare our approach to attention-less structured representation learning and to direct attentional pooling~\cite{Attpool}, and then to the state of the art on each dataset. To be consistent with prior work~\cite{mfafvnet,bcnn}, we report the average accuracy per class on MIT-Indoor and the average per image accuracy on the fine-grained datasets.

To evaluate the benefits of our attention-aware feature aggregation framework, we compare it with counterparts that do not rely on attention. In particular, we report results with VLAD pooling, as discussed in Section~\ref{sec:Method}, but also with BoW representations, which can easily be obtained by using the soft assignments to form histograms. 
%\MS{What about global average pooling?} \KN{standard GAP did not converge in few cases, I will add the results soon. GAP with attention values are close to Standalone attentional pooling results } \MS{So maybe we don't need it.} 
To further evidence the benefits of using structured representations, we compare our results with those of the direct attentional pooling strategy of~\cite{Attpool}, which relies on a global average pooling of the attention masks. The results of this comparison for all datasets are reported in Table~\ref{tab:Att_Results}, where we also show the accuracy of the standard VGG-16 model, with fully connected layers transformed into convolutional ones followed by global average pooling. 
%\MS{So maybe we should rather call this VGG-16 in the table? I changed it.} 
Note that our Attentional-NetVLAD outperforms the baselines in all cases, both when using and not using bounding boxes for fine-grained recognition. Note also that using attention consistently helps improving the results,
% (even when comparing direct attentional pooling with FC   \KN{(FC pooling is better in 2 out of 7 cases here, we can skip the sentence comparing FC pooling with attention pooling )}), 
thus showing the importance of reasoning at the level of local features rather than combining information from the entire image in these challenging recognition tasks. 

\begin{figure}[!t]
	\centering
	\hspace{-0.35cm}
	\hspace*{0.8cm}\emph{casino} \hspace*{1.2cm} \emph{studio music}  \hspace*{0.7cm} \emph{operating room}  \hspace*{0.8cm} \emph{video store}  
	\includegraphics[width=12cm]{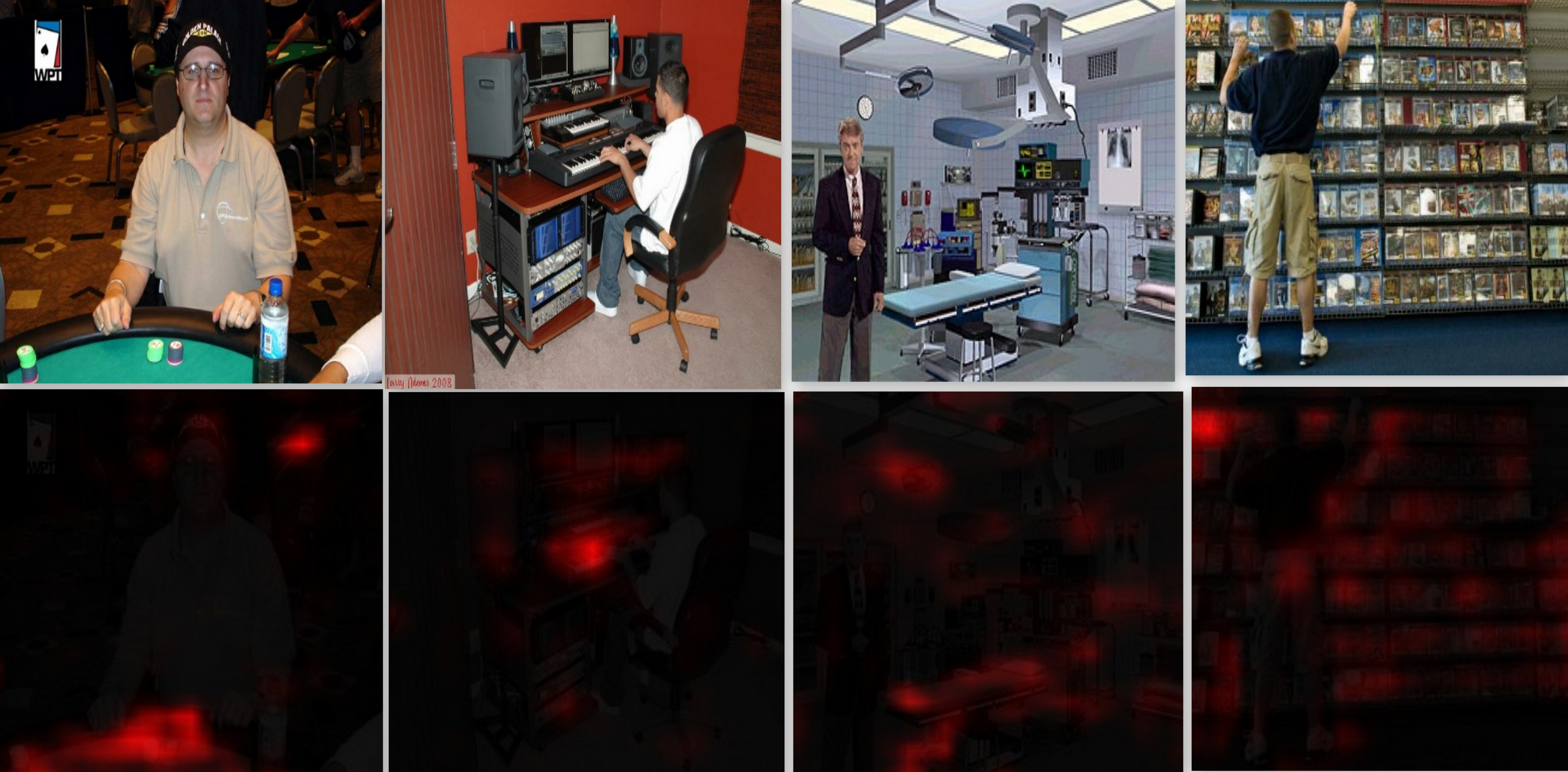}    
	\vspace{0.75cm}
	\caption{\label{fig:MIT_Masks} {\bf Attention maps for MIT-Indoor.} Each column shows an image from a different class (indicated above the image). Note that the maps focus on regions indicative of the label, ignoring the regions common to multiple classes, such as the people.}
	%\vspace{-0.3cm}
\end{figure}

\begin{figure}[!t]
	\centering
	%\hspace{-0.2cm}
	\includegraphics[width=12cm]{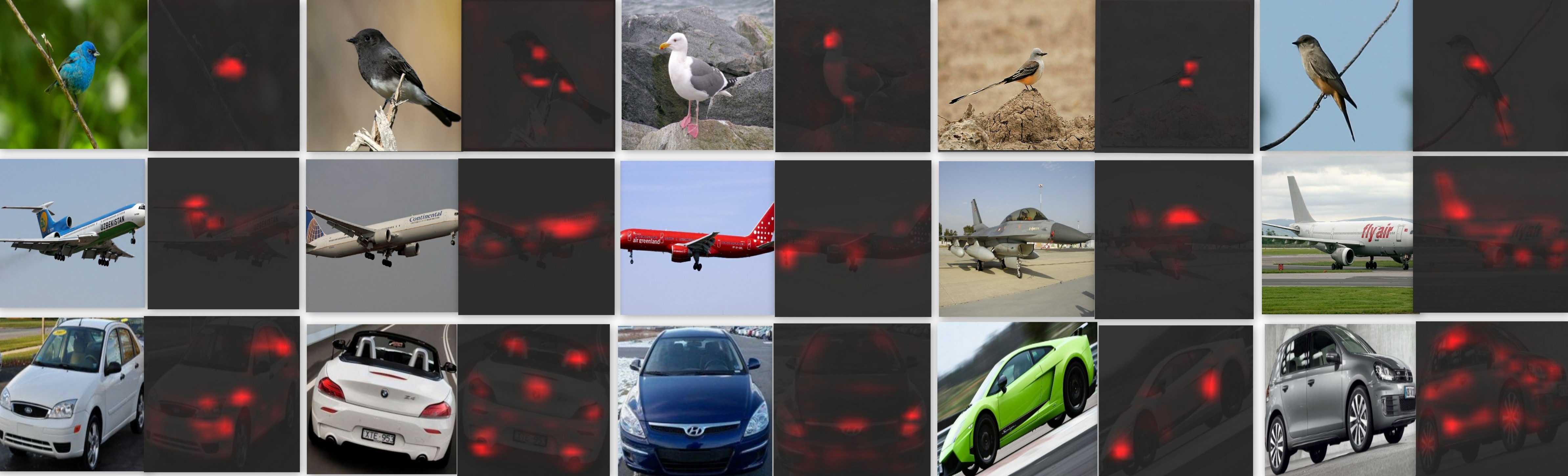}    
	%\vspace{-0.75cm}
	\vspace{0.5cm}
	\caption{\label{fig:finegrained_results} {\bf Attention maps for fine-grained datasets}. Our method is able to localize discriminative parts of birds (tail, beak), aircrafts (engine, landing gear) and cars (lights, logo).}
	%\vspace{-0.7cm}
\end{figure}

In Figs.~\ref{fig:MIT_Masks} and~\ref{fig:finegrained_results}, we provide some representative qualitative results of the attention maps obtained with our method for MIT-Indoor and the fine-grained datasets, respectively. For scene recognition, note that our network learnt to focus on the discriminative regions, such as the casino table and the piano, while ignoring regions shared by other classes, such as people. Similarly, for fine-grained categorization, the network is able to locate discriminative parts, such as the beak and the tail of birds, the brand logo and the head lights of cars, and the engine and landing gears of airplanes. This clearly evidences that our model can, in a single pass, find the regions of interest that define a class. 
%\MS{Have you tried to compare our attention maps with those obtained by attentional pooling~\cite{Attpool}?} \MS{Can we show typical failure cases, maybe in supplementary material?} \KN{the heatmaps from attention pooling~\cite{Attpool} are visually same as the maps from final joint model. Failure cases are mostly from high inter-class similarity at local level, mainly  bookstore-library, livingroom-bedroom etc. }

\begin{table}
	\small
	\centering
	\begin{tabular}{|c|c|c|c|c|c|}
		\hline
		Pooling             & Anno.      & Birds & Cars & Aircrafts& MIT-Indoor\\
		\hline
		VGG-16                  & BBox       & 79.9 & 88.4 & 86.9 & -  \\         
		Attention           & BBox       & 77.2 & 90.3 & 85.0 & -  \\   
		NetBoW              & BBox       & 74.4 & 89.1 & 85.6 & -  \\     
		Attentional-NetBoW  & BBox       & 80.5 & 91.2 & 89.3 & -  \\     
		NetVLAD             & BBox       & 82.4 & 89.8 & 88.0 & -  \\     
		Attentional-NetVLAD & BBox       & \textbf{85.5} & \textbf{93.5} & \textbf{89.2}    & - \\
		\hline
		VGG-16                  &            & 76.0 & 82.8 & 82.3    & 76.6 \\         
		Attention           &            & 77.0 & 87.4 & 81.4    & 77.2 \\   
		NetBoW              &            & 68.9 & 85.2 & 79.9  & 76.1 \\     
		Attentional-NetBoW  &            & 76.9 & 90.6 & 88.3  & 76.6 \\     
		NetVLAD             &            & 80.6 & 89.4 & 86.4  & 79.2 \\     
		Attentional-NetVLAD &            &  \textbf{84.3} & \textbf{92.8} & \textbf{88.8} & \textbf{81.2}\\   	
		\hline
		
	\end{tabular}
	\vspace{0.1in}
	\caption{\label{tab:Att_Results} Comparison of our attentional structured pooling scheme with attention-less (VGG-16, NetBoW, NetVLAD) and structure-less (Attention) baselines. Our approach consistently outperforms these baselines, thus showing the benefits of pooling only the relevant local features into a structured representation.}
	%\vspace{-0.8in}
\end{table}

%\vspace*{1cm}
\comment{
\subsection{Attention-aware structured representations}
\label{sec:Att_structured}

In this part, we experiment our proposed attention-aware structured representation model and report the results in Table~\ref{tab:Att_Results}. For MIT-Indoor dataset, We observe a significant boost in accuracy  with NetVLAD representation with attention. This is due to suppressing of non-relevant regions and only encoding informative regions. As also shown in Figure~\ref{fig:MIT_Masks}, we pick images with human presence across 4 different classes, we observe that attention is given to regions specific to the label. For e.g., the attention module focuses only the casino table in column 1,  piano area in column 2, ignoring human regions, which is not discriminative.

On the fine-grained datasets, we observed attention-aware system to perform better than attention-less system. We show the most discriminative parts of images in Figure~\ref{fig:finegrained_results}. Clearly, the attention module was able to locate fine-grained details  like tail, beak area on birds dataset. Same is the case with cars dataset with stron attention to tyre, head light, symbol areas. On the flight dataset, we observed the attention was focused on the engine areas, tail region, landing gears etc.  By focusing on these small discriminative parts, the VLAD feature encodes a powerful 
}

Finally, we compare our results with the state of the art on each individual dataset. These comparisons are provided in Table~\ref{table:mit_comparison} for MIT-Indoor and Table~\ref{table:comp_finegrained} for the fine-grained datasets. In the case of scene recognition, we outperform all the baselines, including MFAFVNet~\cite{mfafvnet}, which relies on an accurate Fisher Vector encoding of 500K dimensions based on multi-scale image patches. For fine-grained recognition, we outperform all the baselines when relying on bounding boxes. Without bounding boxes, we achieve accuracies only slightly lower than the state-of-the-art methods, such as~\cite{RACNN2017,MACNN2017}, which were tailored to the fine-grained categorization problem, and rely on a multi-stage approach involving cropping parts and processing them separately. By contrast, our approach makes use of a single forward pass through a network and generalizes to any complex recognition scenario.

\comment{
\subsection{Comparison to State of art}
\label{sec:state-of-the-art}

\paragraph{MIT-Indoor Dataset:} We compare our method with the current state-of-art techniques in Table~\ref{table:mit_comparison}. Deep FisherNet~\cite{deepfishernet} formulates an approximate fisher vector encoding with multi-scale patch extraction achieving $76.5\%$ accuracy.
B-CNN~\cite{bcnn} method with large 250k dimension feature vector to pool the features from two-stream network and achieves $79.5\%$.
Our method has achieved the state-of-the-art of $81.40\%$ outperforming current state-of-the-art MFAFVNet~\cite{mfafvnet} of 81.12\%, which uses accurate fisher vector encoding of 500K dimension with multi-scale patch approach. 
}

\def \hfillx {\hspace*{-\textwidth} \hfill}

\begin{table}
	\begin{minipage}{0.3\textwidth}
	\centering
	\small	
	\begin{tabular}{|c|c|}
		\hline
		Method                                & Avg. Acc.\\
		\hline
		Deep FisherNet~\cite{deepfishernet}   & 76.5  \\
		CBN~\cite{compatcBN}                  & 77.6  \\
		NetVLAD~\cite{netvlad}                & 79.1  \\
		H-Sparse~\cite{sparsecoding}          & 79.5  \\
		B-CNN~\cite{bcnn}                     & 79.5  \\
		SMSO~\cite{Secondorderkaicheng2018}   & 79.7  \\
		FV+FC~\cite{FVFC}                     & 81.0   \\
		MFAFVNet~\cite{mfafvnet}              & 81.1  \\
		\hline
		\textbf{Ours}                  & \textbf{81.2}  \\
		\hline
	\end{tabular}
	%\vspace{-0.05cm}
	\vspace{0.1in}
	\caption{\label{table:mit_comparison}Comparison with the state of the art on MIT-Indoor.}
	\end{minipage}
	\hfill
	\begin{minipage}{0.6\textwidth}
	\centering
	\small	

	\begin{tabular}{|r|r|c|c|c|c|c|}
		\hline
		Method                            &   Anno. &  Birds & Cars & Aircraft \\
		\hline
		MG-CNN~\cite{MG_CNN2015}         &  BBox      & 83.0 & - & 86.6 \\
		B-CNN~\cite{bcnn}                &  BBox   & 85.1 & - & - \\
		PA-CNN~\cite{PACNN2015}          &  BBox   & 82.8 & 92.8 & - \\
		Mask-CNN~\cite{maskcnn}          &  Parts & 85.4 & - & - \\
		MDTP~\cite{MDTP}             &  BBox & - & 92.6 & 88.4 \\
		\textbf{Ours}                        & BBox  & \textbf{85.5} & \textbf{93.5} & \textbf{89.2}   \\      
		\hline
		KP~\cite{KP2017}                     &             & 86.2 & 92.4 & 86.9 \\
		Boost-CNN~\cite{boostedCNN}  	     &  & 86.2 & 92.1 & 88.5 \\
		B-CNN~\cite{bcnn}  	                 &  & 84.1 & 86.9    & 86.6    \\
		Imp. B-CNN~\cite{improvedbcnn}   &  & 85.8 & 92.0 & 88.5 \\
		RA-CNN~\cite{RACNN2017}  	         &  & 84.1 & 92.5 & 88.2 \\
		MA-CNN~\cite{MACNN2017}	             &  & \textbf{86.5} & \textbf{92.8} &  \textbf{89.9} \\
		\textbf{Ours}                        &  & 84.3 & \textbf{92.8} & 88.8\\      
		\hline
	\end{tabular}
	\vspace{0.1in}
	\caption{\label{table:comp_finegrained}Comparison with the state of the art on fine-grained datasets.}
\end{minipage}
\end{table}

\comment{
\paragraph{Finegrained-Indoor Dataset:}

We compare with two types of baselines based on annotations of  bounding box(bbox) or parts in Table!\ref{table:comp_finegrained}

On data with bounding box annotations, our method has achieved state-of-art on Standford Cars better than PA-CNN~\cite{PACNN2015} which uses co-segmentation technique to align the image and finds nearest neighbours on extracted CNN-features.
On Birds dataset with annotations, our method performs in comparable to Mask CNN~\cite{maskcnn} which uses part annotations to train separate part detectors and further aggregate the part featutres to train a big classifier.  This method indirectly removes the background descriptors by using only part annotations, whereas our method automatically suppresses the regions of non importance.

On the annotation-free experiments, our methods has performed competitively on cars and flight dataset. We achieved 92.6\% on cars dataset close to MA-CNN\cite{MACNN2017} which use multi-attention network to estimate $k$ informative regions and uses cropped input to pass through CNN multiple times. Other methods include ~\cite{RACNN2017}, a recurrent network which looks at cropped regions attended by higher scales. While ~\cite{RACNN2017,MACNN2017} passes inputs mutliple times(cropped attention images) through the base network, our model is simple with a single feed-forward pass and a very light architecture.
}

% !TEX root = bmvc_review.tex
% !TEX spellcheck = en-US
\vspace{-0.2cm}
\section{Conclusion}
\label{sec:Conclusion}
\vspace{-0.2cm}

We have introduced an attention-aware structured representation network for complex visual recognition tasks. Our network jointly identifies the informative image regions and learns a structured representation. Our comprehensive experiments on scene recognition and fine-grained categorization have demonstrated the superiority of our approach over attention-less strategies. Our approach is general and can be extended to other feature aggregation techniques, or can make use of any generic attention module. This will be the focus of our future work.

% !TEX root = bmvc_review.tex
% !TEX spellcheck = en-US

\newpage
%\section{References}
\label{sec:References}

\bibliography{egbib}
% !TEX root = bmvc_review.tex
% !TEX spellcheck = en-US

\newpage
\section{Supplementary Results}

In this section, we provide additional experiments and qualitative results to further support those in the main paper.

\vspace{-0.35cm}
\subsection{Influence of $\lambda$ on Classification Accuracy}
\vspace{-0.2cm}
We first evaluate the influence of the hyper-parameter $\lambda$ in Eq.~7 of the main paper, which defines the strength of the attention module loss, on the classification accuracy. To this end, we evaluate our approach for different values of $\lambda$ on the CUB-200 bird dataset, after 20 training epochs and without performing any data augmentation (image flipping) at inference time. The results of this experiment are provided in Table~\ref{tab:lamda_results}. Note that accuracy is stable over a very large range of values, thus showing that our approach is robust to the choice of this parameter. %We found, however, that the model does typically not converge for values greater than 1.

\begin{table}[!h]
\vspace{-0.4cm}
	\small
	\centering
	\begin{tabular}{|c|c|c|c|c|}
		\hline
		$\lambda$             & 0.0001 & 0.01  & 0.4 & 1 \\
		\hline
		Accuracy          &  83.3   & 83.7 & 83.7 & 83.1 \\  
		\hline

	\end{tabular}
	\vspace{0.1in}
	\caption{\label{tab:lamda_results}Influence of $\lambda$ on the final classification accuracy.}
	%\vspace{-0.8in}
\end{table}

\vspace{-0.6cm}
\subsection{Attentional Global Average Pooling}
\vspace{-0.2cm}
While our main goal was to introduce an attention mechanism in structured representations, our approach also applies to unstructured pooling strategies, such as global average pooling (GAP). To illustrate this, we implemented an attentional GAP layer using our attention map. As shown in Table~\ref{tab:GAP_results}, this also typically outperforms the standard GAP strategy, thus further showing the benefits of attention when performing feature aggregation. Note, however, that our attentional VLAD strategy still significantly outperforms the GAP one.

\begin{table}[h]
	\small
	\centering
	\begin{tabular}{|c|c|c|c|c|c|}
		\hline
		Pooling             & Anno.      & Birds & Cars & Aircrafts\\
		\hline
		GAP              & BBox       & 79.8 & 89.3 & 86.6  \\     
		Attentional-GAP  & BBox       & 76.3 & 91.1 & 88.3   \\
				NetVLAD             & BBox       & 82.4 & 89.8 & 88.0   \\     
		Attentional-NetVLAD  & BBox   & 85.5 & 93.5 &89.2    \\          
			\hline
		GAP              &            & 78.6 & 86.2 & 84.5  \\     
		Attentional-GAP  &            & 77.8 & 89.6 & 85.5   \\  
		NetVLAD             &            & 80.6 & 89.4 & 86.4  \\     
		Attentional-NetVLAD  &        & 84.3 & 92.8 & 88.8   \\        
		\hline
		
	\end{tabular}
	\vspace{0.1in}
	\caption{\label{tab:GAP_results} Attentional Global Average Pooling on fine-grained datasets.}
	%\vspace{-0.8in}
\end{table}

\newpage
\subsection{Additional Qualitative Results}
Below, we show the attention maps obtained with our approach for additional randomly-sampled images from the four datasets used in the main paper.

\begin{figure}[!ht]
	\centering
	%\vspace{-0.3in}
	\includegraphics[width=0.95\linewidth, height=17cm]{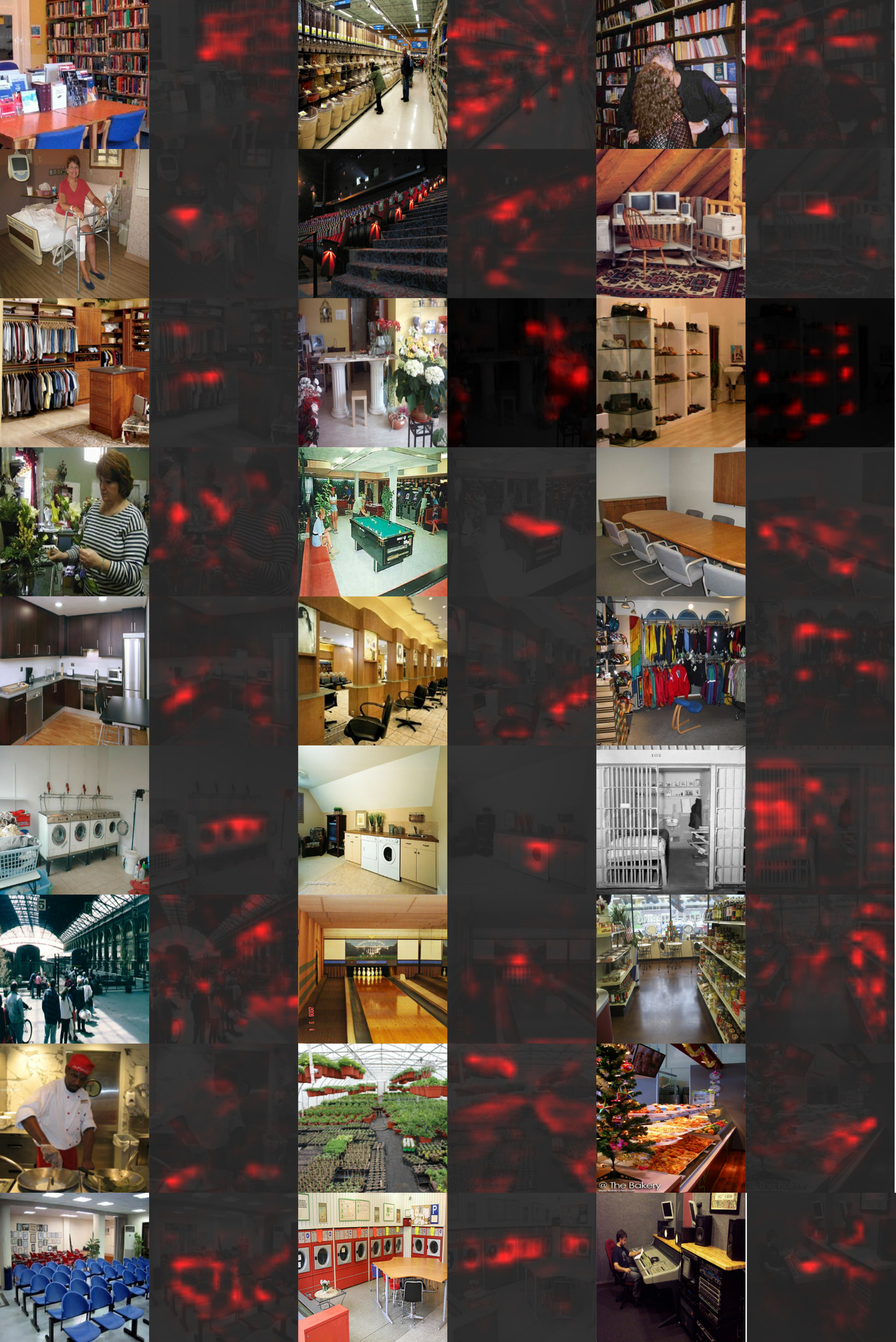}
	\vspace{0.2in}
	
	\caption{\label{fig:arch} Generated attention maps on the MIT-Indoor dataset.}
	%\vspace{-0.1in}
	%{-0.2in}
\end{figure}
\newpage

\begin{figure}[!ht]
	\centering
	%\vspace{-0.3in}
	\includegraphics[width=0.95\linewidth]{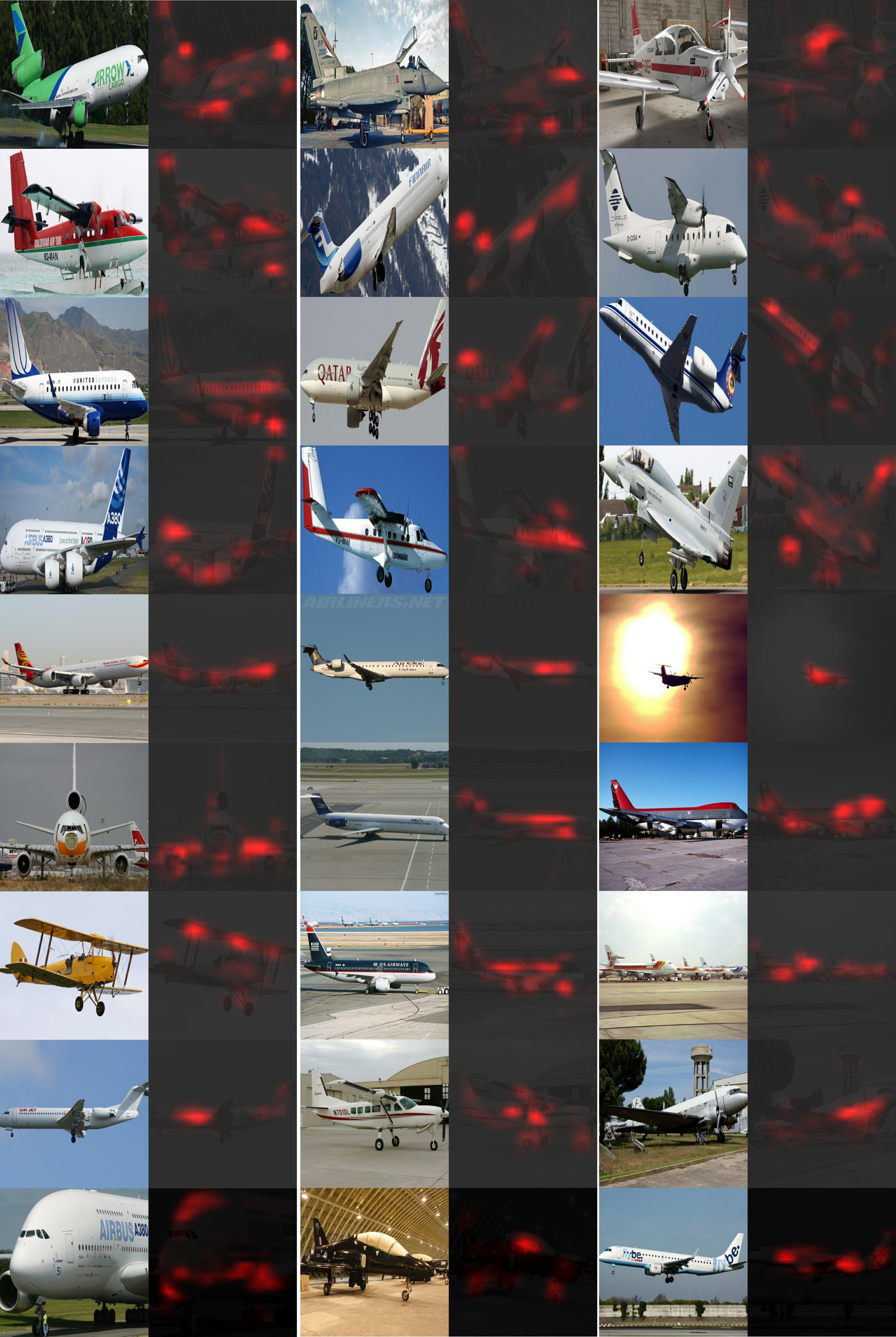}
	\vspace{0.2in}
	
	\caption{\label{fig:Air} Generated attention maps on the Aircrafts dataset.}
	%\vspace{-0.1in}
	%{-0.2in}
\end{figure}

\newpage

\begin{figure}[!ht]
	\centering
	%\vspace{-0.3in}
	\includegraphics[width=0.95\linewidth]{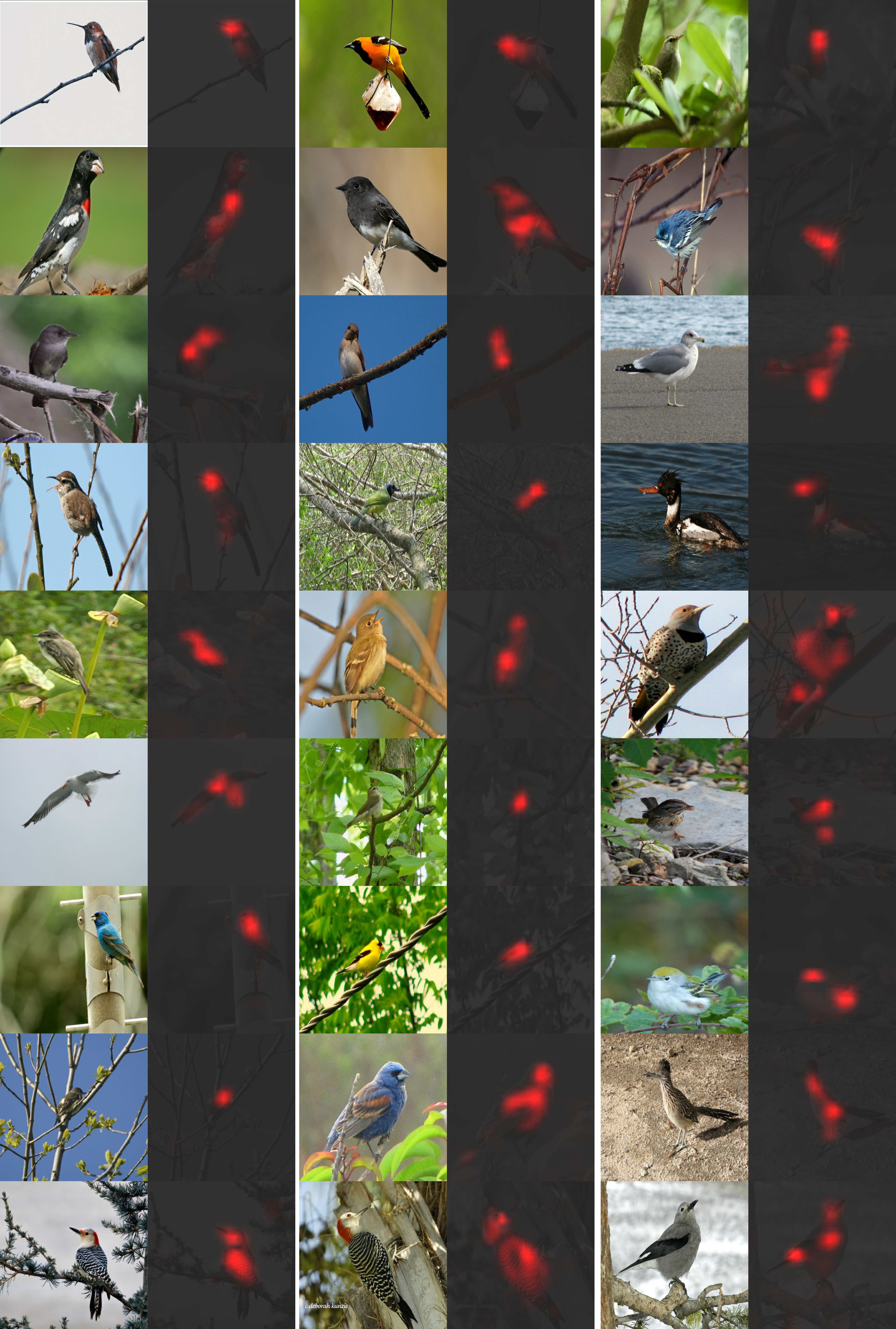}
	\vspace{0.2in}
	
	\caption{\label{fig:Birds} Generated attention maps on the Birds dataset.}
	\vspace{-0.1in}
	%{-0.2in}
\end{figure}

\newpage

\begin{figure}[!ht]
	\centering
	%\vspace{-0.3in}
	\includegraphics[width=0.95\linewidth]{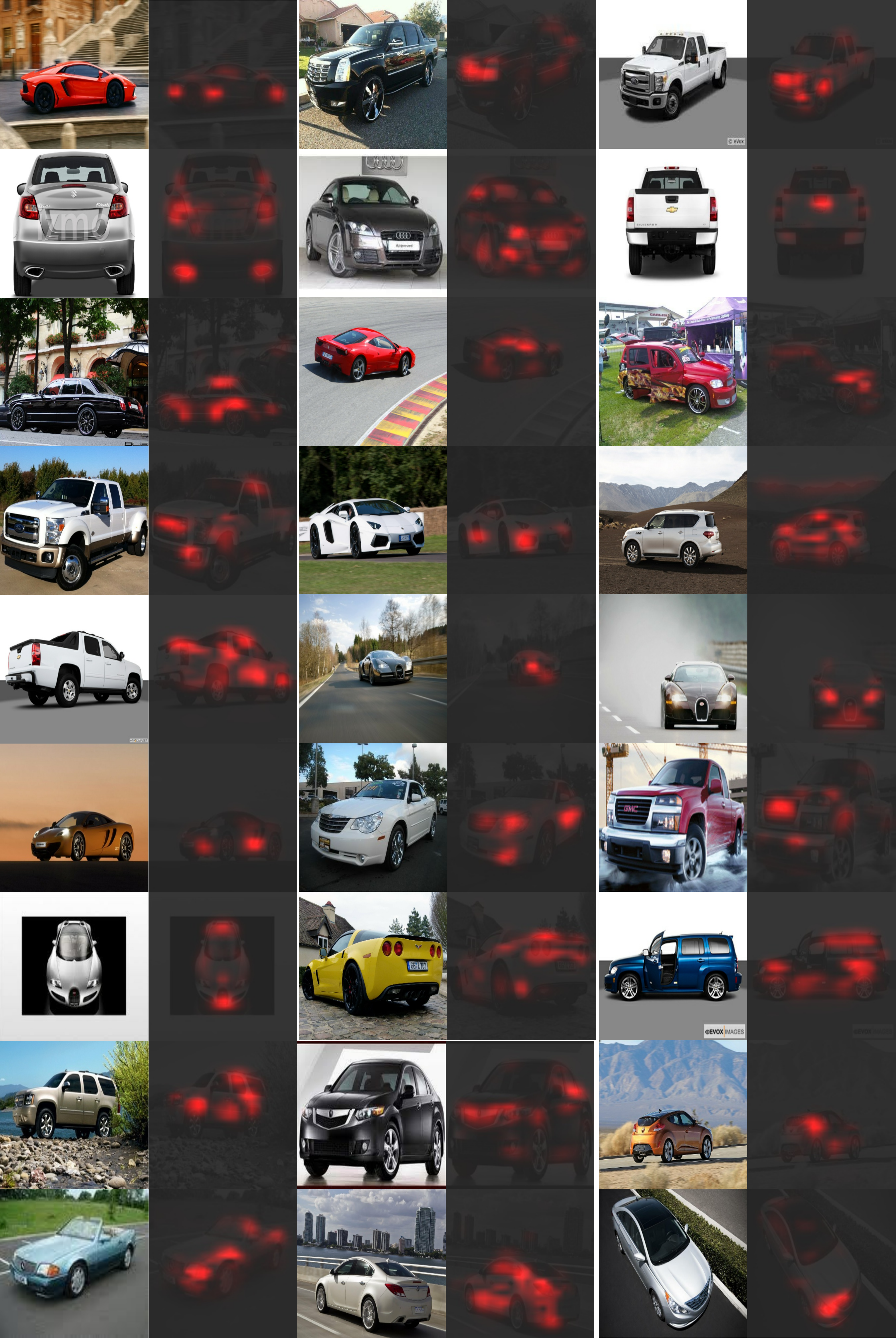}
	\vspace{0.2in}
	
	\caption{\label{fig:SCars} Generated attention maps on the Stanford-Cars dataset.}
	\vspace{-0.1in}
	%{-0.2in}
\end{figure}

\newpage
\newpage

\subsection{Failure Cases}
\begin{figure}[!ht]
	\centering
	%\vspace{-0.3in}
	\includegraphics[width=0.95\linewidth]{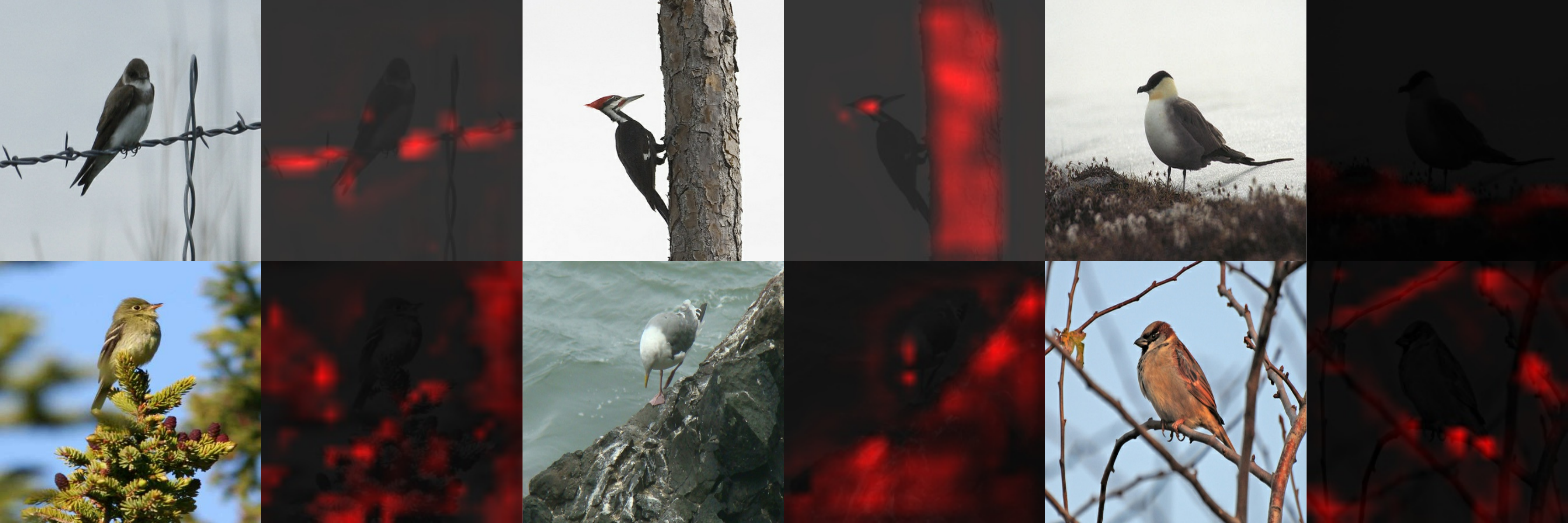}
	\vspace{0.2in}
	
	\caption{\label{fig:failure} Failure cases of our model particularly due to incorrect attention.}
	\vspace{-0.1in}
	%{-0.2in}
\end{figure}

Finally, in Fig.~\ref{fig:failure}, we show some typical failure cases of our approach, such as attention to background regions on Birds dataset.

\end{document}